# Turbofan Engine Remaining Useful Life (RUL) Prediction Based on Bi-Directional Long Short-Term Memory (BLSTM)

Abedin Sherifi*
*Pratt and Whitney*
*December 2024*

***Abstract*—The aviation industry is rapidly evolving, driven by advancements in technology. Turbofan engines used in commercial aerospace are very complex systems. The majority of turbofan engine components are susceptible to degradation over the life of their operation. Turbofan engine degradation has an impact to engine performance, operability, and reliability. Predicting accurate remaining useful life (RUL) of a commercial turbofan engine based on a variety of complex sensor data is of paramount importance for the safety of the passengers, safety of flight, and for cost effective operations. That is why it is essential for turbofan engines to be monitored, controlled, and maintained. RUL predictions can either come from model-based or data-based approaches. The model-based approach can be very expensive due to the complexity of the mathematical models and the deep expertise that is required in the domain of physical systems. The data-based approach is more frequently used nowadays thanks to the high computational complexity of computers, the advancements in Machine Learning (ML) models, and advancements in sensors. This paper is going to be focused on Bi-Directional Long Short-Term Memory (BLSTM) models but will also provide a benchmark of several RUL prediction data-based models. The proposed RUL prediction models are going to be evaluated based on engine failure prediction benchmark dataset Commercial Modular Aero-Propulsion System Simulation (CMAPSS). The CMAPSS dataset is from NASA which contains turbofan engine run to failure events.**

## I. Introduction

The recent technology trends have led to innovative changes for aircraft maintenance and reliability. Critical aircraft components require accurate failure and predictive maintenance to avoid major aircraft on ground (AOG) interruptions. Engines play a critical role for the aviation industry, and are the most important pieces of equipment on an airplane. Engines are some of the most complex components that are made up of thousands of components that drive for very accurate maintenance. Engine maintenance can be classified in one of the following three categories:

1) **Corrective Maintenance** - Maintenance to address issues that occur during operation. These maintenance procedures are not planned.
2) **Preventive Maintenance** - Maintenance that is done per scheduled basis irrespective of component health.
3) **Predictive Maintenance** - Maintenance that is done per anticipation of potential failures. This maintenance takes place before the failures occur.

Per [1], the value of engine maintenance, repair, and overhaul (MRO) for the year 2016 was valued at $27 billion and the value for the year 2026 is supposed to reach $40 billion. Per [2] turbofan engine failures contribute to 60% of all airplane failures. Reasons such as safety of people on-board the plane and costs associated with AOGs are main reasons of why having a very accurate RUL prediction is of utmost importance. The RUL of a turbofan engine is the time in cycle left for the turbofan engine to perform its intended functions without any failures. Based on the literature review, RUL predictions are classified into two categories: model-based and data-based. The model-based methods use mathematical and physical models to make predictions. Some of the famous model-based approaches for RUL predictions per [3] are the Corrosion Model, Abrasion Model, Taylor Tool Wear Model, Hidden Markov Models, etc. The model-based methods require very deep understanding of the system at hand and have very high computational complexity. On the other hand, data-based methods use the data history of the system to make RUL predictions. Some of the most widely used data-based methods for RUL predictions per [3] are the following: Regression Analysis, Fuzzy Logic, Gradient Boosting (GB), Neural Networks, Random Forrest, Support Vector Machines (SVM), Relevance Vector Machines (RVM), etc.

## II. Problem Description

Maintenance techniques such as predictive, reactive, and preventive are used for managing engine health, reducing aircraft on ground time, and avoiding large losses. Turbofan engines are the most critical parts on a plane; hence, it is of great importance to perform fault estimation and remaining useful life predictions to prevent catastrophic failures. Effective management of data is very important due to the complexity of accurately predicting remaining useful life as mentioned in [4]. This effort is part of the Prognostics and Health Management (PHM) system. While a number of studies have demonstrated a variety of neural network model architec-

* A. Sherifi is with the Artificial Intelligence Department, University of Texas at Austin. A. Sherifi is with Pratt and Whitney, a RTX Company.

tures for RUL predictions as some are mentioned in [5], none of them go in depth required specific to the implementation of BLSTM with numerous changes to hyperparameters.

## III. Literature Review

Most approaches for RUL predictions fall in two categories: model-based and data-based approach. The model-based approach is based on physics and mathematical models such as Kalman filters or Particle filter per [6] which require a deep expertise of the turbo-machinery and physics. The research by [7] worked on the Relevance Vector Machine (RVM) to predict good health probability of components in a engine. In [8], the paper demonstrated the use of XGBoost algorithm towards feature importance and the use of Temporal Convolution Network (TCN) for the RUL predictions. The team in [1] used ensemble learning algorithms which combined Random Forests, Recurrent Neural Networks (RNNs), AutoRegressive (AR) models for RUL predictions by optimizing the weights for the outputs from each model. In [9] the team used the Stacked Spare Autoencoder (SAE) to make RUL predictions with a combination of grid search algorithm to automatically select hyperparameters for the SAE algorithm. In [10] the team proposed a hybrid architecture based on 1D-CNN-LSTM where the 1D-CNN (Convolution Neural Network) architecture was used to get spatial information from sensors and LSTM architecture was used to get temporal information from the features.

## IV. Methodology

### *A. Dataset Description*

The turbofan engine is a gas turbine engine that is most widely used in aviation industry. The turbofan engine consists mainly of the following parts as explained in detail by [11] : Fan, Low Pressure Compressor (LPC), High Pressure Compressor (HPC), Combustor, High Pressure Turbine (HPT), Low Pressure Turbine (LPT), and Exhaust. The air from the Fan enters the LPC and then HPC where it gets compressed to very high pressure. This high compressed air is mixed with the fuel in the Combustor and is burned. The Combustor generates very high-velocity gasses that pass through the turbines and out via the exhaust which in return propel the plane forward as explained in [12].

CMAPPS is a simulation framework developed by NASA for examining engine degradation. It is highly important to understand some of the engine components used on this simulation dataset. The turbofan engine is a thermal engine that consists of many thermo-mechanical components, and the following are some of the sensor values used in the dataset: Low Pressure Compressor (LPC) pressure, High Pressure Compressor (HPC) pressure, High Pressure Turbine (HPT) temperature, Low Pressure Turbine (LPT) temperature, pressure sensors for other sections of the engine, temperature sensors for other sections of the engine, etc. The dataset used for this study has the following breakdown per [13]:

1) **train_FD001.txt** - Contains data for 100 engines till engine is failed.
2) **test_FD001.txt** - Cycles for the 100 engines are given but before engine is failed.
3) **RUL_FD001.txt** - RUL values for the 100 engines given in the test data.

The table below provides a summary of all the columns used in the **train_FD001.txt** file:

| **train_FD001.txt** | |
|---|---|
| **Column** | **Dataset Attributes** |
| 1 | Engine Unit Number |
| 2 | Time in Cycles of the Engines |
| 3 | Operational Setting 1 |
| 4 | Operational Setting 2 |
| 5 | Operational Setting 3 |
| 6 - 26 | Sensor Measurements |

TABLE I: Dataset Attributes

Figure 1 depicted below captures the mean for each feature in the dataset called train_FD001.txt. This figure provides a better understanding of the overall distribution and impact of features in the specified dataset.

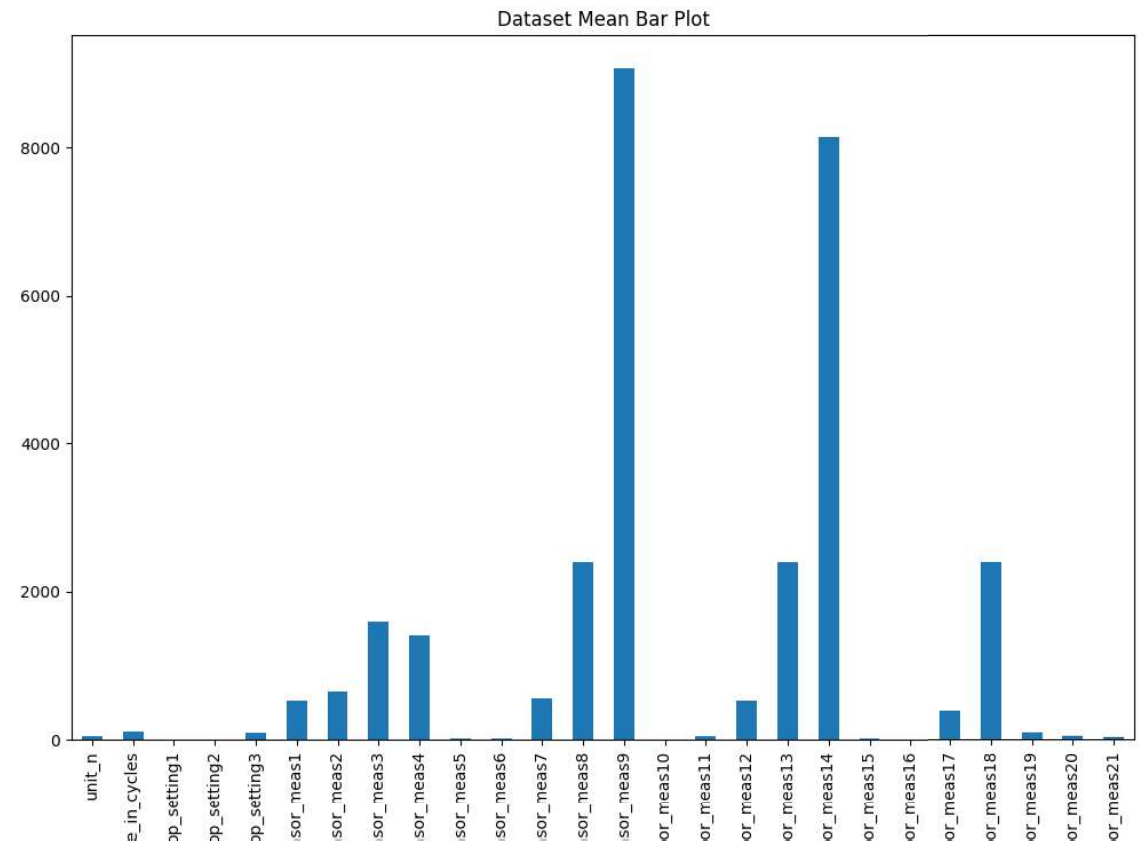

Fig. 1: Mean of All Features in the Validation Set

Another plot that will provide even more insight into the dataset is a plot that shows the operational cycles for the different engine units before a fault occurs. Figure 2 shown below provides the aforementioned summary. From the figure below we are able to see that for the majority of the engine units, the number of operational cycles to a fault event is between 200 and 250. The minimum number of cycles to a fault event is at 128, the average number of cycles to a fault event is at 206, and the maximum number of cycles to a fault event is at 362.

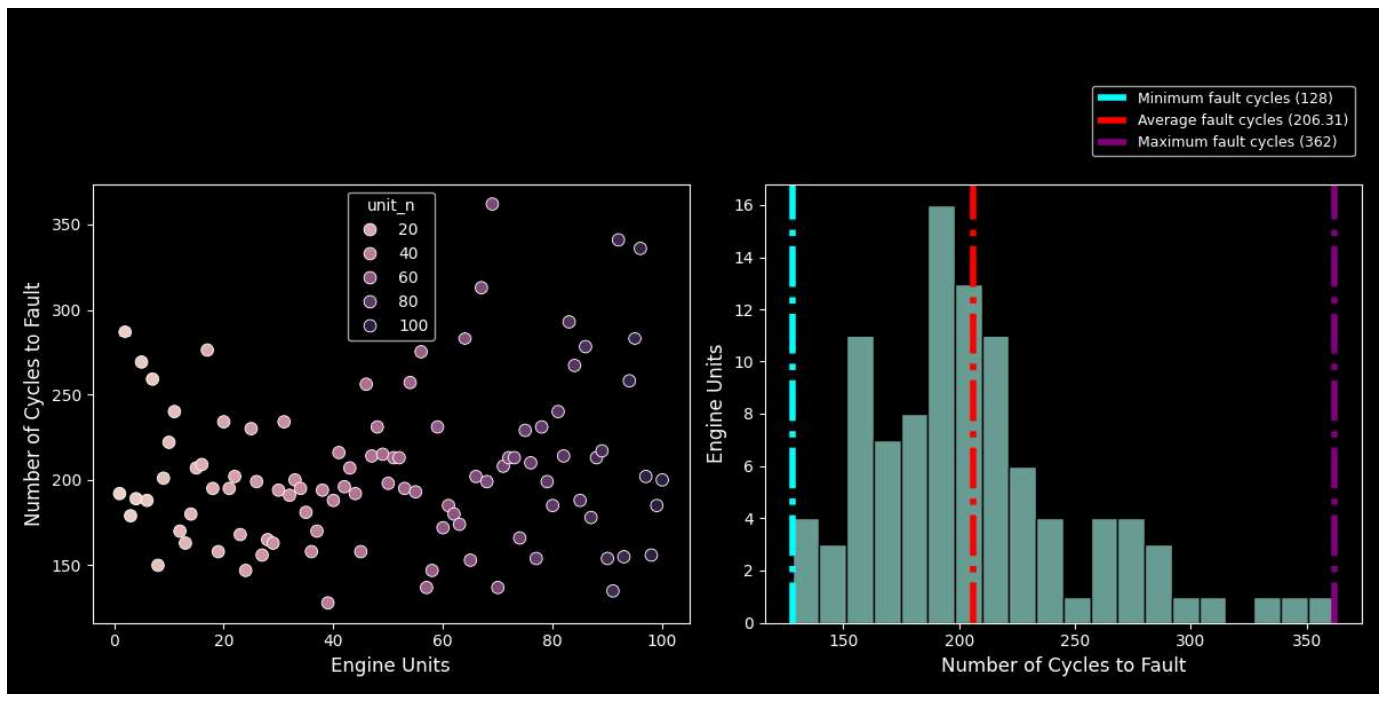

Fig. 2: Number of Cycles Before a Fault

Figure 3 below shows the correlation heatmap for the training dataset. Through the heatmap, we are able to distinguish between highly correlated features or variables and slightly correlated features. Features with correlation value of 0 have been dropped from the dataset since they do not provide much useful information. Some features have very high correlation of almost 1 or -1. Some of these features were also dropped to avoid redundant information in the model.

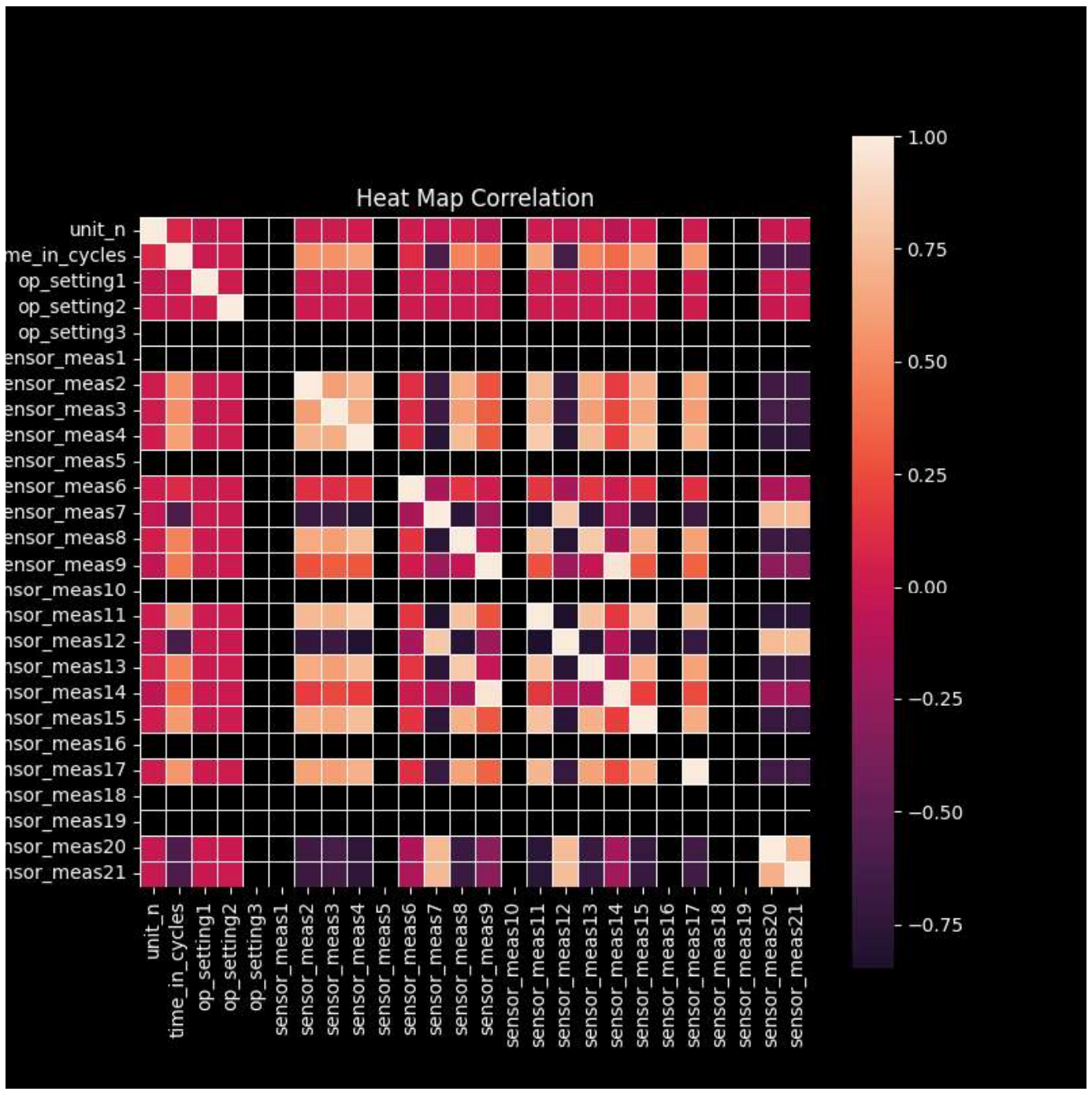

Fig. 3: Correlation Heatmap

Figure 4 shows all the data plots for all the sensors for engine unit 5 only as an example.

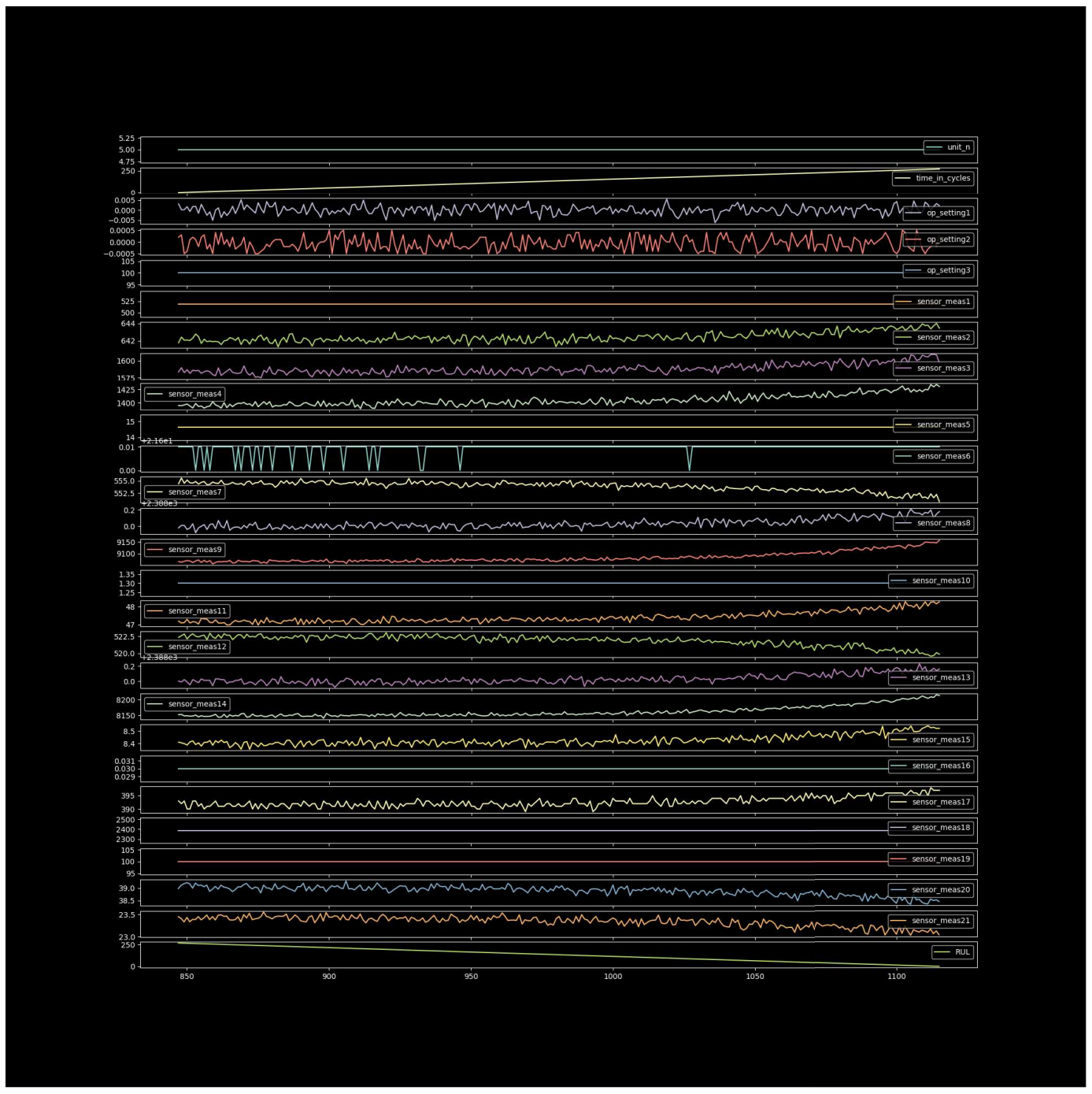

Fig. 4: Engine Unit 5 Data from Sensors

### B. Data Pre-Processing

The dataset train_FD001.txt is considered for validation training. The shape of this dataset is (20631, 26). First, a number of features were removed since they do not provide any valuable information for the model. The first pre-processing was looking at columns that have very little variability or very high correlation based on the correlation heatmap mentioned in the section above. The list below captures columns removed from this assessment:

1) Unit Number
2) Op Setting 1
3) Op Setting 2
4) Sensor Measurement 1
5) Sensor Measurement 5
6) Sensor Measurement 6
7) Sensor Measurement 9
8) Sensor Measurement 10
9) Sensor Measurement 14
10) Sensor Measurement 16
11) Sensor Measurement 18
12) Sensor Measurement 19

The removal of the features above was cross-checked based on the criteria presented in the formula below:

$$Feature\,to\,be\,Removed : \sigma \leq 0.005 * \mu\,and\,N_u \leq 5 \quad (1)$$

Where ...
$\sigma - standard \quad deviation \quad of \quad feature$
$\mu - mean \quad of \quad feature$

$N_u - number \quad of \quad unique \quad values \quad in \quad the \quad feature$

A new feature called RUL is added to the validation dataset based on the number of cycles the engine units have gone through. Before any additional processing, the unit number and time in cycles features are also dropped from the validation set. The next step is the normalization of the dataset to a range of [0,1] using the Min-Max scaling. A power transformation is also applied to the scaled dataset to make the data more Gaussian normally distributed.

The Min-Max scaling is a normalization technique based on the following formula as mentioned in [14]:

$$x' = \frac{x - min(X)}{max(X) - min(X)} \tag{2}$$

Where ...
$x'$ is the scaled value
$x$ is the original value
$min(X)$ is the minimum value of the feature in the dataset
$max(X)$ is the maximum value of the feature in the dataset

Figure 5 depicted below captures all the steps that were followed for pre-processing all the way up before model construction.

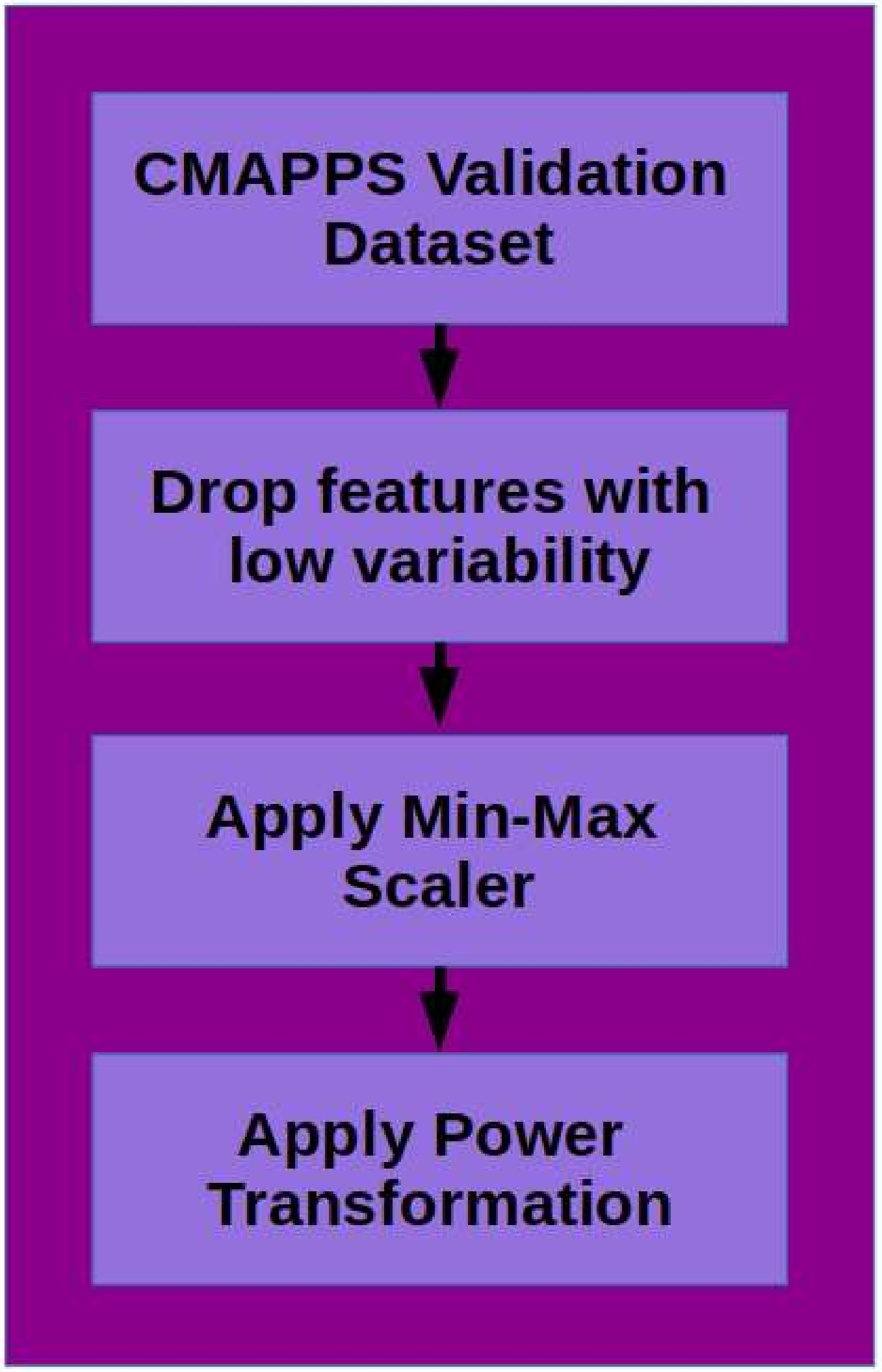

Fig. 5: Dataset Pre-Processing

Principal Component Analysis (PCA) was also run for some of the model training iterations. PCA was also applied to the overall dataset. However, that application was problematic since it disregards the temporal dependencies in the data especially for RUL predictions. PCA was applied on a per-time step basis for the individual features, however, that alone did not result in any high accuracies for the model in making RUL predictions. Figure 6 captures the PCA Biplot as an example even though it is not something that was used for this project due to its short-backs in temporal data.

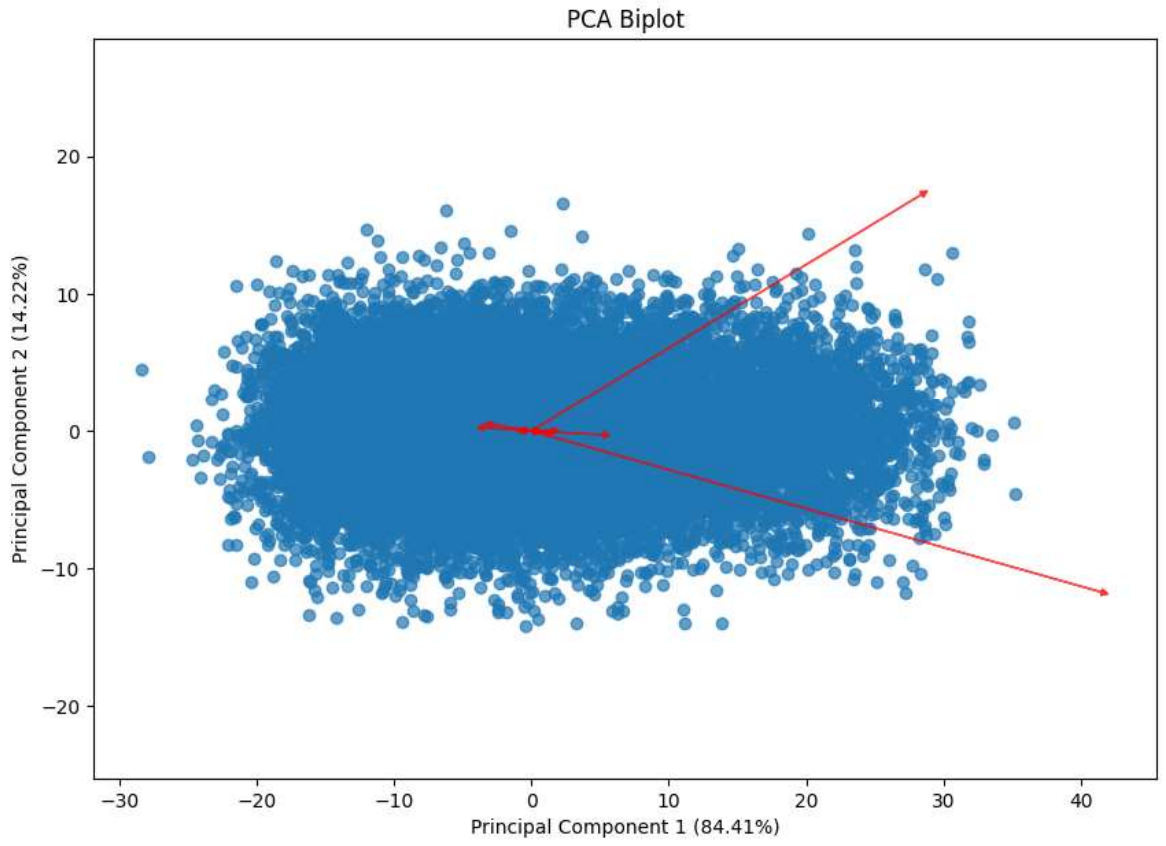

Fig. 6: PCA Biplot

### C. *Neural Network (NN) Model Architecture*

The NN architecture that will be discussed in detail in this section is the NN model that performed best. That model is the Bi-Directional Long Short-Term Memory (BLSTM). BLSTM model is a form of Recurrent Neural Network (RNN). RNNs allow data to hold through the entire model over time since the output is fed back to the input of each layer. Figure 7 below from [15] captures this representation.

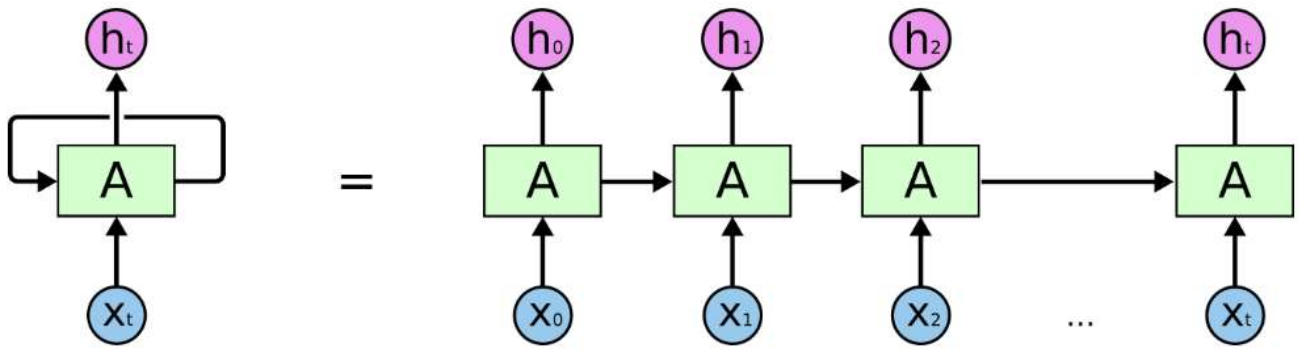

Fig. 7: RNN Structure [15]

LSTM structure is best for relating data for long period of times. The LSTM module consists of the following parts:

1) Cell State - allows data to transfer through the model selectively based on the utilization of gates.
2) Forget Gate - this gate makes the decision on what data to select from the cell state.
3) Input Gate - this gate makes the decision on what new data to feed to the cell state.
4) Output Gate - this gate decides what the hidden state should for a specific time step.

Figure 8 below from [15] captures the LSTM module structure:

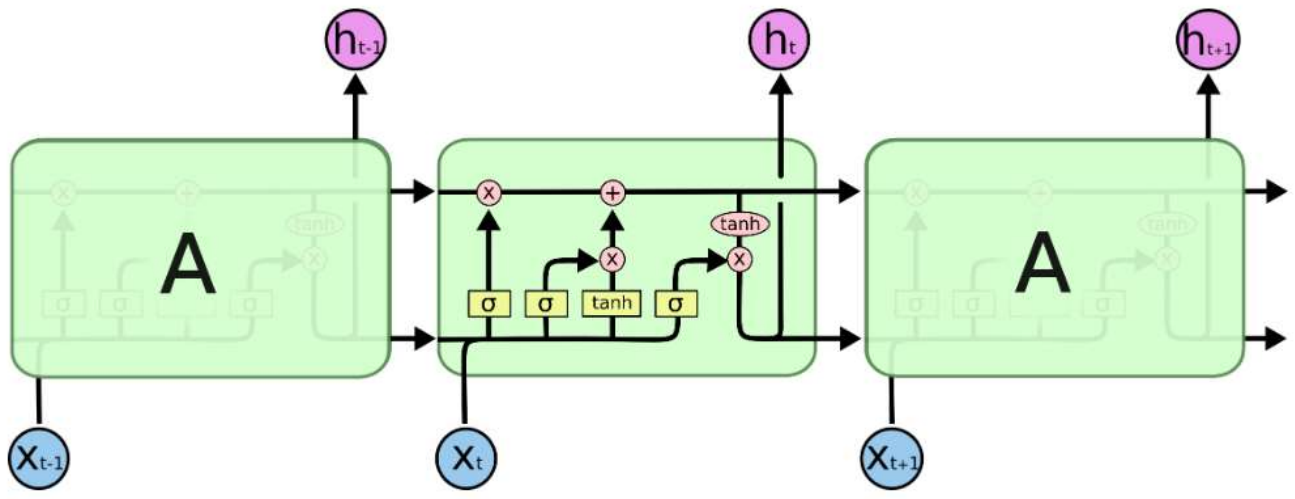

Fig. 8: LSTM Structure [15]

LSTM processes information in one direction such as forward direction, and BLSTM processes information in two directions such as forward and backward direction. BLSTM uses two LSTM layers where one layer is used for the forward direction and the other LSTM layer is used in the backward direction. Further technical overview of BLSTM architecture is provided by [16]. A BLSTM structure is depicted in the figure 9 below from [17]:

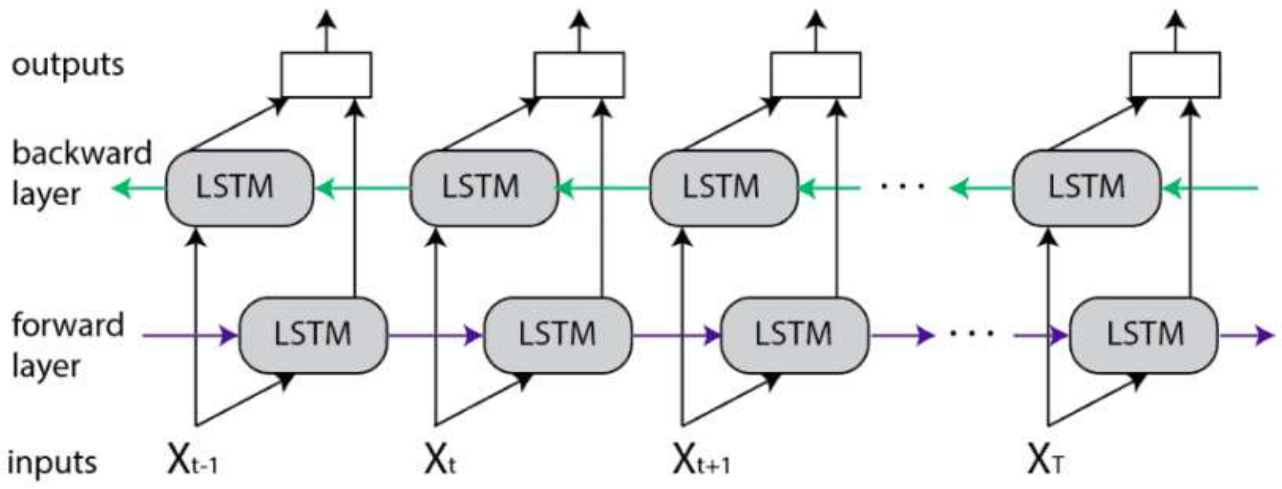

Fig. 9: BLSTM Structure [17]

BLSTM is more powerful that LSTM for sequential data where history for example of sensor data changes is paramount for RUL predictions.

Figure 10 below captures the BLSTM based model architecture that was selected as the best performer for this project.

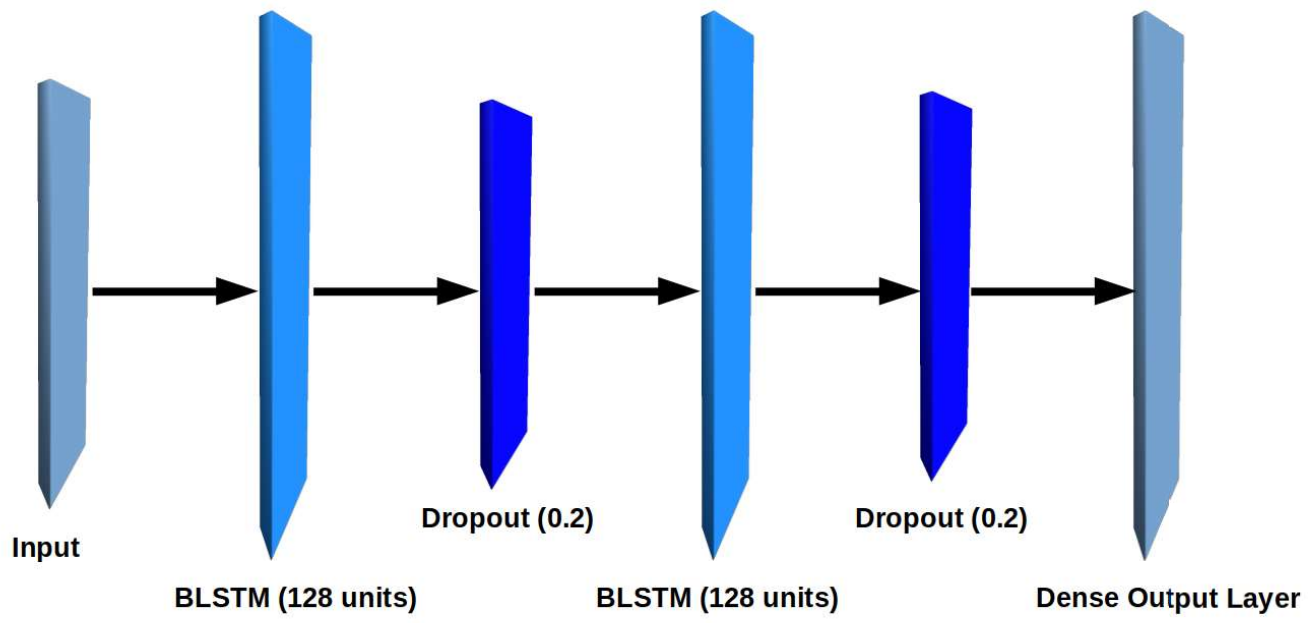

Fig. 10: BLSTM Model Architecture

### D. *Model Training*

Figure 11 below captures the overall architectural steps for the NN model implementation.

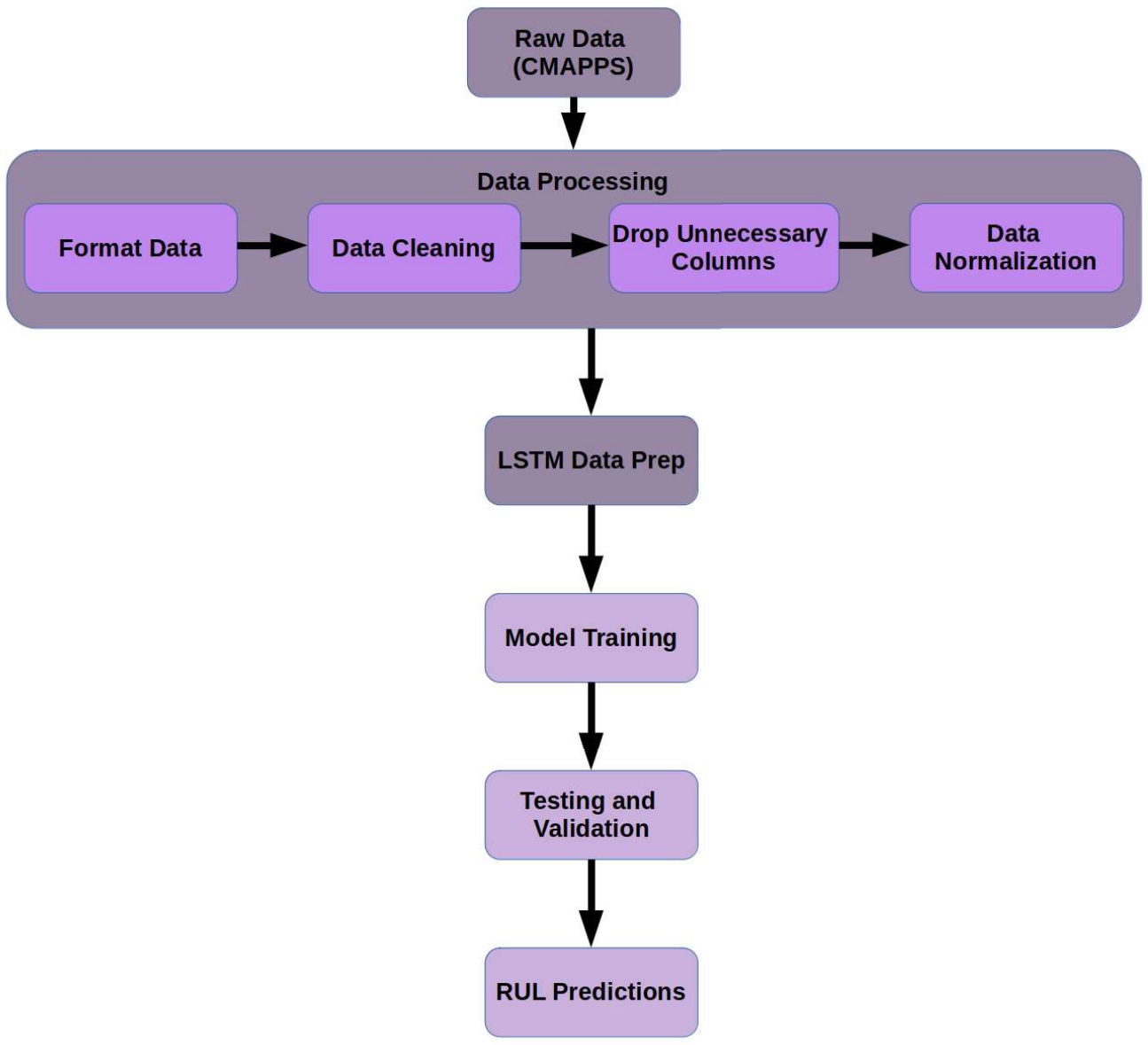

Fig. 11: Overall Architectural Steps

The validation dataset was split between the x_train data and the y_train data. The shape of x_train data was set to (20631, 12) where 12 represents the 12 features used in the model. The shape of y_train data was set to (20631, 1) where 1 here represents the target which is the computed RUL based on the validation dataset. Several models were used and results were bench-marked against one another. The first sequential model involved the following layers:

- LSTM layer with 128 units with input shape = (number of time steps, number of features)
- Dense layer with 1 unit as the final layer for the output

The baseline model is compiled with a loss function of Mean Squared Error (MSE) and an Adam optimizer. LSTM is based on a Recurrent Neural Network (RNN) which learns from sequences of data; hence, it is very well-suited for the RUL predictions which are based on time-series data. LSTMs use gates such as input gate, forget gate, and output gate to control the flow of information as needed.

Another model was constructed based on Bi-directional LSTM (BLSTM) with the following characteristics:

- BLSTM layer with 128 units with input shape = (number of time steps, number of features)
- Dense layer with 1 unit as the final layer for the output

Another model was constructed based on BLSTM but with dropout layers. Here are the layers in more detail:

- BLSTM layer with 128 units with input shape = (number of time steps, number of features)
- Dropout layer of 0.2 implying 20% probability
- BLSTM layer with 128 units
- Dropout layer of 0.2 implying 20% probability
- Dense layer with 1 unit as the final layer for the output

Finally, used a BLSTM-based architecture with 3 BLSTM layers, 3 dropout layers, 2 dense layers, and 2 batch normalization layers. Details are given below:

- BLSTM layer with 512 units with input shape = (number of time steps, number of features)
- Dropout layer of 0.4 implying 40% probability
- Batch normalization layer
- BLSTM layer with 256 units
- Dropout layer of 0.4 implying 40% probability
- Batch normalization layer
- BLSTM layer with 128 units
- Dropout layer of 0.4 implying 40% probability
- Dense layer with 11 units
- Dense layer with 1 unit as the final layer for the output

The BLSTM is supposed to make the model learn from the past and the future of the input sequences. A linear regression model was also used for benchmark purposes. Lastly, a model based on BLSTM was used with batch normalization to improve convergence during training. In total 5 different models were run with varying hyperparameters which will be captured in more details in the next section.

Some of the other techniques that were employed with some of the models above were early stopping and automatic reduction of learning rate. The learning rate reduction was implemented such that if the validation loss stopped improving for a number of epochs, then the learning rate would reduce by some factor. On the other hand, early stopping was implemented such that if validation loss was not improved for a number of epochs, then the training would stop and the model would revert back at using the best weights from the training.

### *E. Hyperparameter Tuning*

Hyperparameter tuning is a methodology or standard work that allows for the optimization of a model performance. Hyperparameters are configured before the model is trained. The list below captures the main hyperparameters that were touched in this project:

- Learning Rate - Adjustment of weights depending on error gradient
- Batch Size - Number of training samples in one iteration of training
- Epochs - The number of times the training dataset is run through the model
- Hidden Layer Units - The number of neurons per layer
- Dropout Layer - Percent of neurons to randomly dropout to avoid overfitting
- Batch Normalization Layer - During training the input to the layer is normalized by making sure that it has a mean of 0 and a variance of 1

The specific characteristics of the hyperparameter tuning are explained in detail per [18] and [19]. From the list above, all of the hyperparameters had an impact to the model performance except batch normalization which did not deem as critical for the BLSTM model performance.

## V. RESULTS

Results were compiled together in a table based on a set of error metrics. Specifically the following metrics were used to review results for the RUL predictions from all the models used:

- Mean Squared Error (MSE) - The average between the squared differences between prediction and actual values. Smaller values for MSE indicate that model predictions are closer to true values.

$$MSE = \frac{1}{n}\sum_{i=1}^{n}(y_i - \hat{y}_i)^2$$

where n is the total number of observations, $y_i$ is the actual value and $\hat{y}_i$ is the prediction.

- Root Mean Squared Error (RMSE) - Square root of MSE. Smaller values for RMSE indicate that model predictions are closer to true values.

$$RMSE = \sqrt{\frac{1}{n}\sum_{i=1}^{n}(y_i - \hat{y}_i)^2}$$

where n is the total number of observations, $y_i$ is the actual value and $\hat{y}_i$ is the prediction.

- Mean Absolute Error (MAE) - Measure of the average of the absolute difference between the predictions and true values.

$$MAE = \frac{1}{n}\sum_{i=1}^{n}|y_i - \hat{y}_i|$$

where n is the total number of observations, $y_i$ is the actual value and $\hat{y}_i$ is the prediction.

- $R^2$ Score - How well the model explains variances in the target variable. $R^2$ of 1 represents perfect predictions.

$$R^2 = 1 - \frac{\sum_{i=1}^{n}(y_i - \hat{y}_i)^2}{\sum_{i=1}^{n}(y_i - \bar{y})^2}$$

where n is the total number of observations, $y_i$ is the actual value, $\hat{y}_i$ is the prediction, and $\bar{y}$ is mean of the actual values.

The described error metrics are explained in detail in [20]. The error metrics depict how accurately the model is making predictions in comparison to the true values.

Table II below captures the error metric results for the turbofan engine RUL predictions performed on this paper:

| Model | RMSE | MSE | MAE | $R^2$ |
|---|---|---|---|---|
| Linear Regression | 30.5 | 930.27 | 24.42 | 0.46 |
| LSTM | 27.74 | 769.39 | 21.45 | 0.55 |
| BLSTM | 27.38 | 749.49 | 20.99 | 0.57 |
| BLSTM + Dropout | 26.68 | 711.68 | 20.56 | 0.59 |
| BLSTM + Dropout + Normalization | 27.75 | 770.22 | 21.88 | 0.55 |

TABLE II: Comparison of error metrics for 4 models

From the table above, it is obvious that the model with BLSTM and dropout layers (BLSTM + Dropout) performs the best compared to the rest of the models.

Figure 12 depicted below plots the training loss and validation loss for the BLSTM with dropouts model. The training loss is calculated after each epoch on the training data. The validation loss is calculated on the validation data. From the plot we can see that some overfitting is possible since validation loss is increasing and then plateauing.

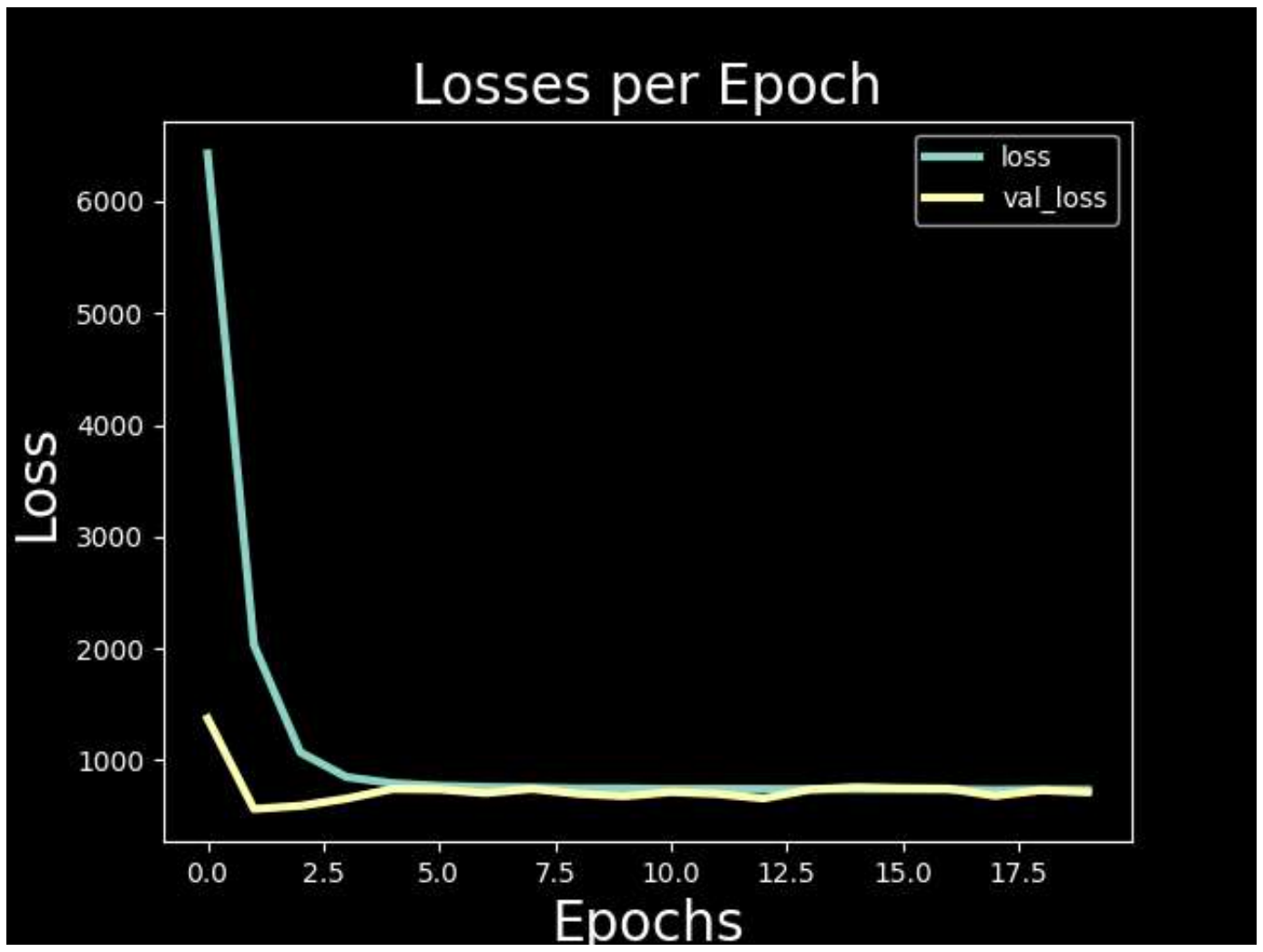

Fig. 12: BLSTM w/Dropot Training and Validation Losses per Epoch

Figure 13 below plots the ideal predictions which are on the redline. The scattered points are the actual predictions which based on the plot below we can see that the predictions have some error. The model is making somewhat accurate predictions, but there is definitely a symptom of overfitting that could be investigated in the future to be corrected.

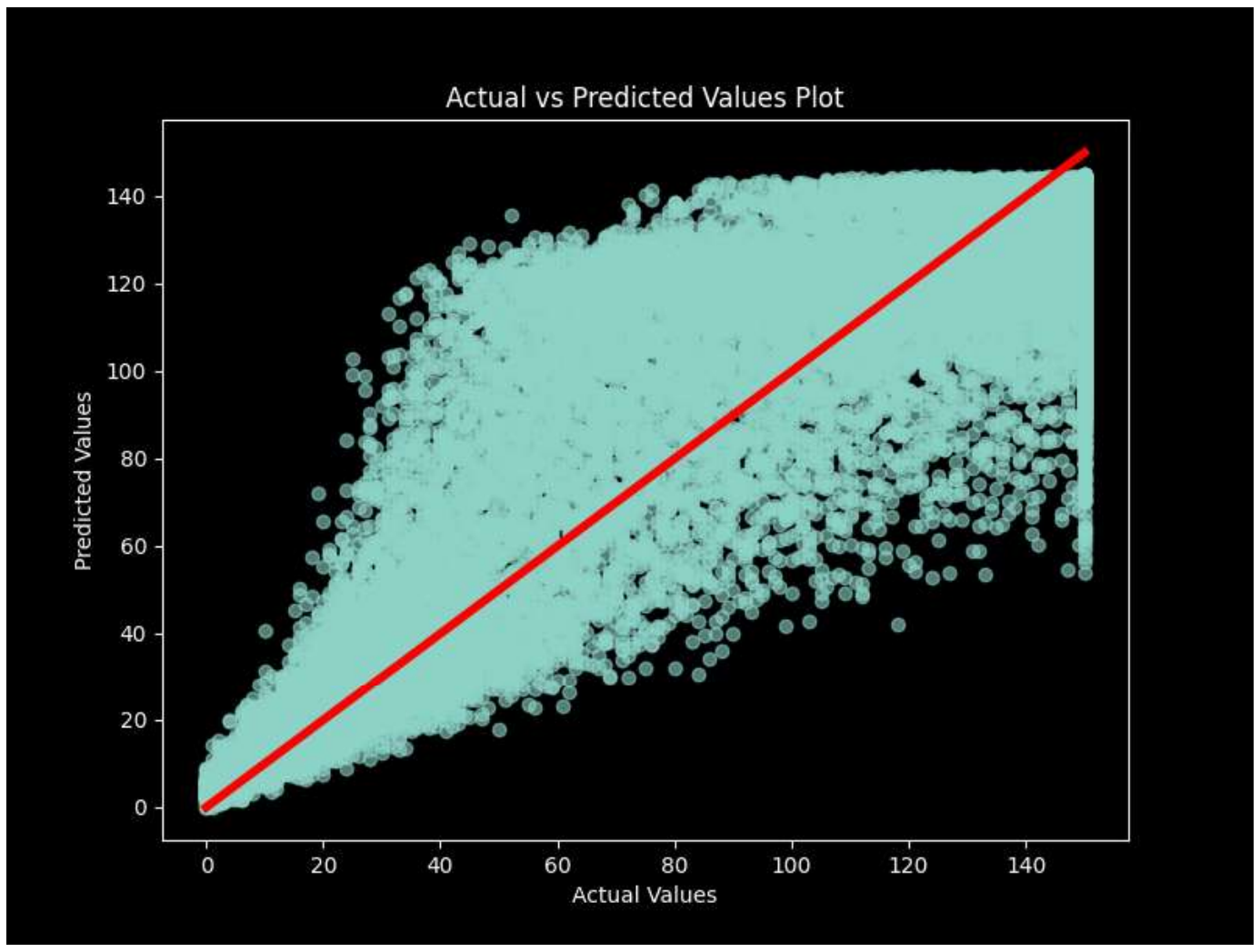

Fig. 13: BLSTM w/Dropout Actual vs. Predicted Values

Figure 14 below captures another representation of model's accuracy at making RUL predictions. Blue points are the model predictions and yellow points are the actual RUL values. The closer the points are together, that would imply good model performance.

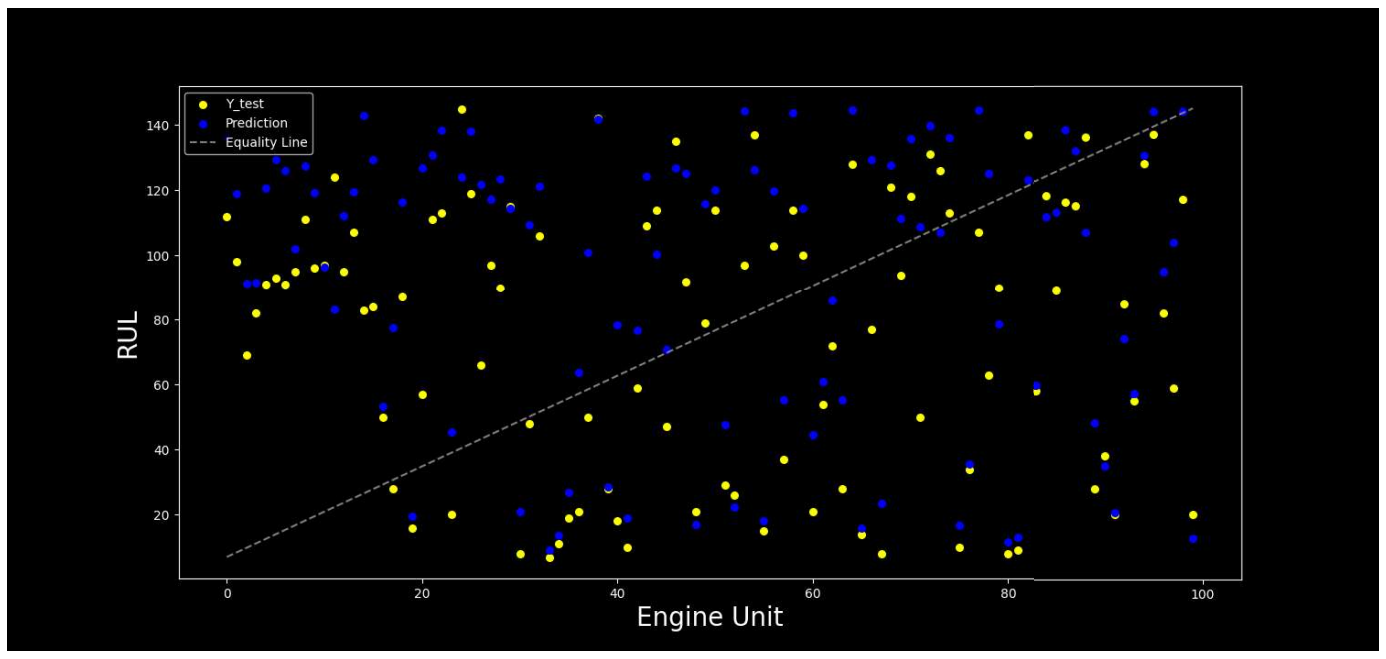

Fig. 14: BLSTM w/Dropout Actual vs. Predicted Values - Scatter Plot

Figure 15 tries to visually represent the differences between predicted RULs and actual RULs in a bar chart. Blue bars are predicted values and yellow bars are actual values. If the bars for a given engine unit are of same height, that would represent perfect accuracy.

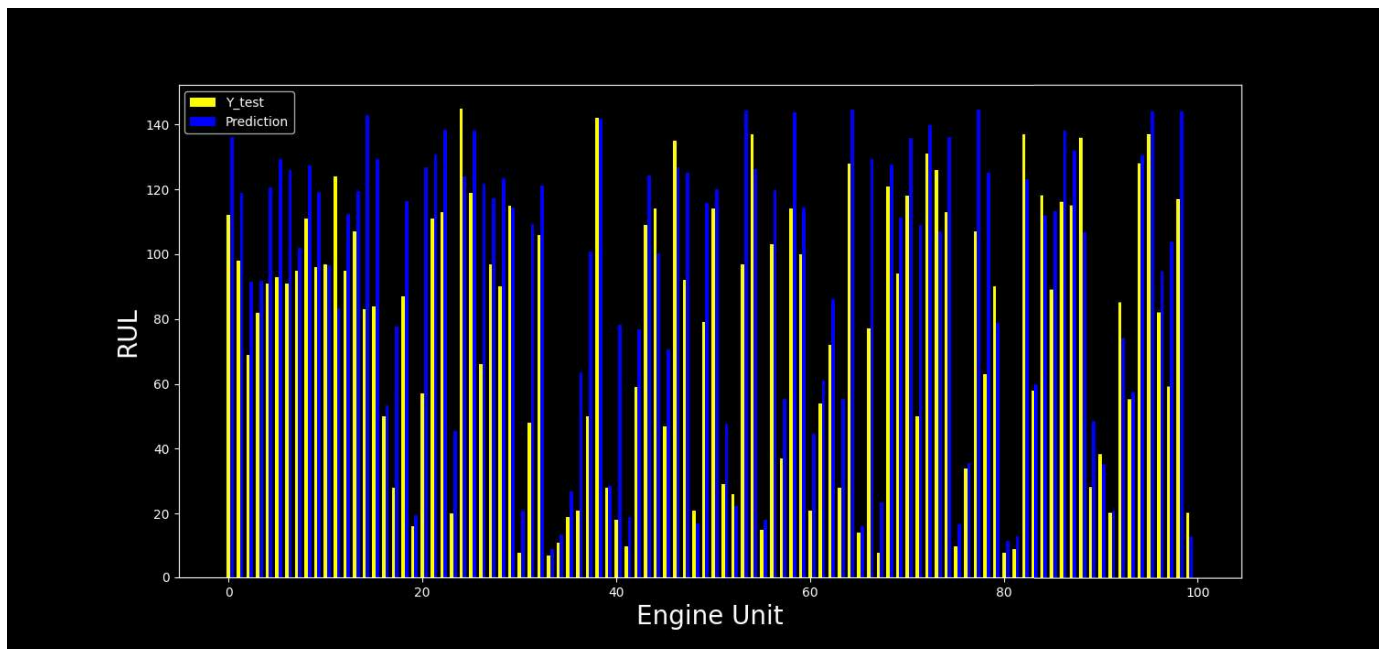

Fig. 15: BLSTM w/Dropout Actual vs. Predicted Values - Bar Plot

Figure 16 depicted below captures the residual plot for the BLSTM with dropouts model. The residual plot is a plot of the prediction errors (actual - prediction). The model below is showing somewhat random residuals which indicate that the model is well performing.

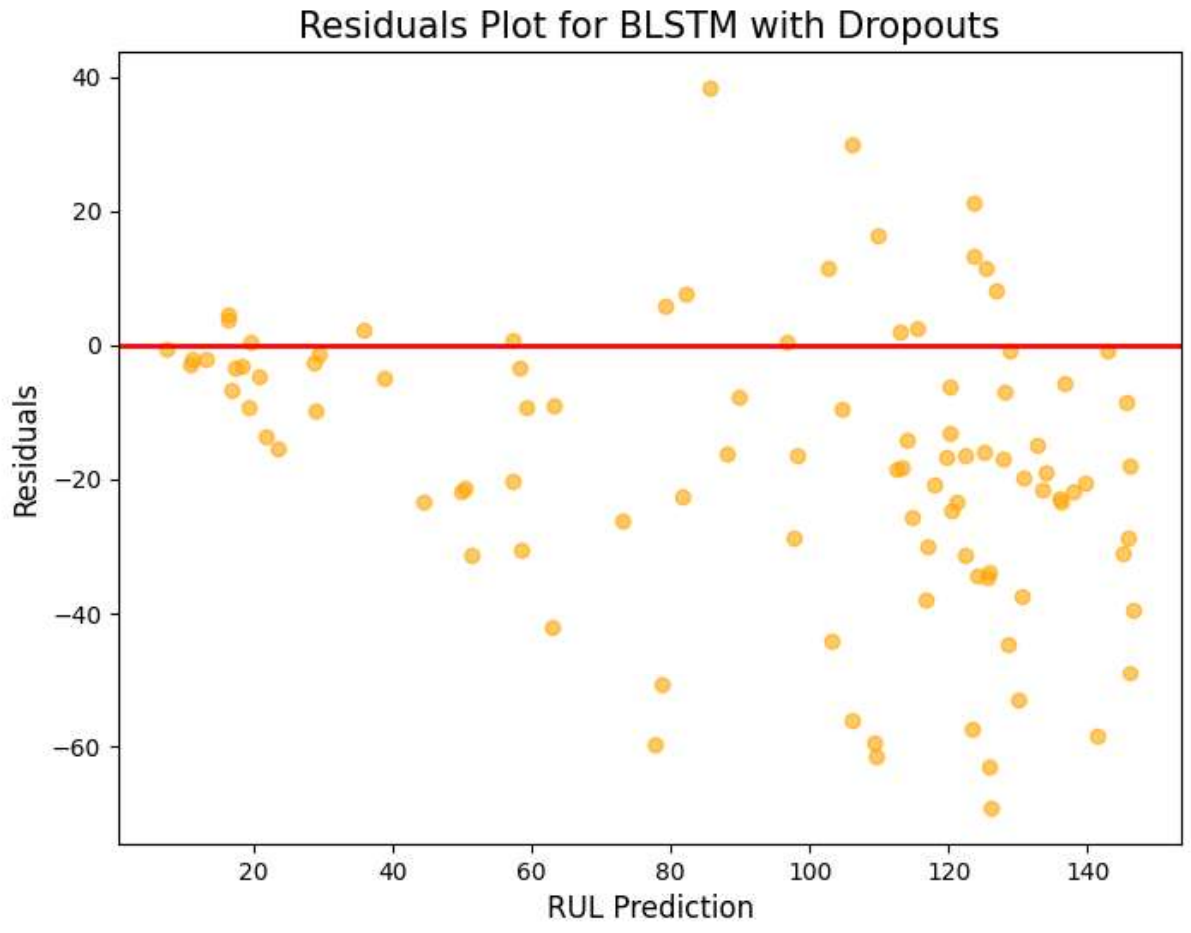

Fig. 16: BLSTM w/Dropout Residuals Plot

## VI. Conclusion

Various neural network learning models for making turbofan engine RUL predictions were explored as part of this study. The models that were discussed are LSTM, BLSTM, and Linear Regression. From all the models that were tested, the BLSTM model with dropouts (BLSTM + Dropout) was the most effective with great performance as demonstrated by the $RMSE$ of 26.68, $MAE$ of 20.56, and $R^2$ of 0.59. These results were able to show that the BLSTM model with dropout regularization captured the sequential dependencies in the turbofan engine RUL predictions while also trying to minimze the effect of overfitting. Deductions can also be made that neural network based models such as BLSTM with dropout regularization performed way better compared to traditional models such as Linear Regression.

There is still room for improvement especially in reducing some of the error metrics even further. Future work could include a focus on additional tuning of hyperparameters and ability to run the model through other datasets.